\pdfoutput=1

\documentclass[11pt]{article}
\usepackage[]{coling}

\usepackage{times}
\usepackage{latexsym}
\usepackage{url}
\usepackage{inconsolata}
\usepackage{amsmath}
\usepackage{algorithm}
\usepackage[noend]{algpseudocode}
\usepackage{graphicx}
\usepackage{tabularx}
\usepackage{adjustbox}
\usepackage{subcaption}
\usepackage{array}
\usepackage{multirow}
\newcolumntype{?}{!{\vrule width 2pt}}

\usepackage[T1]{fontenc}

\usepackage[utf8]{inputenc}

\usepackage{microtype}
\usepackage{lipsum}

\title{Towards Cross-Lingual Audio Abuse Detection in Low-Resource Settings with Few-Shot Learning}

\author{Aditya Narayan Sankaran, Reza Farahbakhsh, Noel Crespi\\
        SAMOVAR, Télécom SudParis\\Institut Polytechnique de Paris
\\91120 Palaiseau, France\\
\texttt{\href{mailto:aditya.sankaran@ip-paris.fr}{aditya.sankaran@ip-paris.fr}}
}

\newcounter{RZNumberOfComments}
\stepcounter{RZNumberOfComments}

\begin{document}
\maketitle
\begin{abstract} \label{abstract}

Online abusive content detection, particularly in low-resource settings and within the audio modality, remains underexplored. We investigate the potential of pre-trained audio representations for detecting abusive language in low-resource languages, in this case, in Indian languages using Few Shot Learning (FSL). Leveraging powerful representations from models such as Wav2Vec and Whisper, we explore cross-lingual abuse detection using the ADIMA dataset with FSL. Our approach integrates these representations within the Model-Agnostic Meta-Learning (MAML) framework to classify abusive language in 10 languages. We experiment with various shot sizes (50-200) evaluating the impact of limited data on performance. Additionally, a feature visualization study was conducted to better understand model behaviour. This study highlights the generalization ability of pre-trained models in low-resource scenarios and offers valuable insights into detecting abusive language in multilingual contexts.

\end{abstract}
\section{Introduction} \label{intro}

The widespread adoption of social media for everyday communication requires safeguards and moderation to create a safe space for the user and the social community. With audio-based social media platforms like Twitter (now X) Spaces, Clubhouse, Discord, ShareChat etc, moderating offensive language and hate speech has become essential in maintaining a safe space for people online. These platforms host users from diverse linguistic backgrounds, especially in multilingual countries like India. India is home to several languages with more than 30 Million speakers of languages and has experienced a phenomenal increase in the use of online social media services, including Facebook, Twitter (now X), Instagram, LinkedIn, and YouTube, with over 250 million users (and growing) helping them in Social interactions and conversations \cite{Ganguly2019The, Palakodety2020Annotation}. 76\% of Indians spend an estimated 1 hour and 29 minutes daily using social media through smartphones, with adolescents between the ages of 13 and 19 making up 31\% of the overall number of people who use social media \cite{Dar2022The, Srivastava2019Social}. 

Given that a large share of users of social media are teenagers and young adults, there is an absolute need to create a safe online space for them to express their views freely without being exposed to hate-filled and offensive content. In an era where social media and entertainment companies are being scrutinized more carefully than ever before, laxity related to user privacy or community rule violations could prove harmful and costly for social audio platforms. There have been reports of incidents where Clubhouse rooms engaged in racist and anti-Semitist talks\footnote{\url{https://tinyurl.com/fslanti}}. This issue raises concerns about how it can be more difficult to safeguard user privacy and security on Audio Social Media platforms as traditional content moderation techniques utilized for text-based media do not necessarily work well with audio-based platforms.

Abusive language detection and Hate Speech detection in Low Resource Language settings has been a popularly researched topic, especially in the text modality. Transfer Learning-based approaches have shown incredible performance in abuse detection using existing popular models like BERT, RoBERTa, XLM-RoBERTa etc \cite{mozafari2020bert, ranasinghe2021}. Works by \citet{mozafari2022cross} and \citet{awal2024} also show the ability of meta-learning-based models to outperform transfer learning-based models in a cross-lingual abuse detection task in Low resource settings. Even abusive content detection on images and videos has been accelerated with the contribution of multimedia datasets and progress in Deep Learning techniques \cite{gao2020offensive, alcantara2020offensive}. Abuse detection in audio is relatively unexplored given the limited availability of such datasets to work with and the complex task of collecting and annotating a large set of audio samples, that too in a variety of languages and accents, especially in highly multilingual countries like India where there are 1369 rationalised languages and more than 500 dialects, with 22 scheduled languages having over 1.17 billion speakers \cite{Khanuja2021MuRIL, Sengupta2018Vision}.

A simple approach for abuse detection is to employ an Automatic Speech Recognition (ASR) model to transcribe the audio to text and then work with them using existing NLP techniques with models like Whisper and Wav2Vec boasting low Word-Error Rates for the transcription task. This is what \citet{ghosh2021detoxy} did by crawling through popular voice datasets for abusive data called DeToxy and performed a Two-Step approach by transcribing audio to text and using BERT for downstream classification. A major problem with these approaches is that they tend to miss out on abusive words since they are usually not spoken clearly and completely, thereby limiting the ability to spot abusive keywords \cite{gupta2022}.

Recent advancements in Pre-Trained Audio Representations, such as those demonstrated by \citet{shor2022} and \citet{saeed2021contrastive}, show promising results in various audio abuse detection-related tasks, particularly embeddings pre-trained on large datasets like Audioset have been successfully transferred to multiple classification tasks, including various audio pattern recognition tasks \cite{kong2020pann}. Building on the success of Meta-Learning techniques, such as Model-Agnostic Meta-Learning (MAML) \cite{finn2017maml}, Prototypical Networks \cite{snell2017prototypical}, and Contrastive Learning \cite{yang2022fewshot}, our work applies MAML in a low-resource setting, leveraging its demonstrated efficacy in tasks such as document classification \cite{vanderheijden2021multilingual}, speech recognition \cite{Singh2022Improved}, and abstract summarization \cite{Huh2022Lightweight}.

Thus, we present the contributions\footnote{Codebase: \url{https://github.com/callmesanfornow/fsl-audio-abuse.git}} of our work enumerated below:
\begin{enumerate}
    \item We propose a MAML-based few-shot cross-lingual audio abuse classification methodology, leveraging pre-trained audio representations from Whisper \cite{radford2022robust} and Wav2Vec \cite{baevski2020wav2vec}. We also assess the effectiveness of these pre-trained features under two feature normalization strategies: L2 normalization and Temporal Mean.
    \item Our method is evaluated on the ADIMA dataset \cite{gupta2022}, with Whisper achieving top accuracy scores ranging from 78.98\% to 85.22\% in the 100-shot setting, using L2-Norm feature normalization.
    \item We provide a visual analysis of pre-trained audio features from the best-performing normalization setting, examining how language similarity can enhance cross-lingual abuse detection, especially in Low-Resource Languages. This study offers valuable insights into optimal strategies for audio abuse detection and identifies potential directions for future research.
\end{enumerate}
\section{Related Works} \label{related}
\subsection{Audio Abuse Detection}

The introduction of DeToxy \cite{ghosh2021detoxy}, a large-scale multi-modal dataset was a significant advancement in the field of audio toxicity detection, particularly in the context of spoken utterances. DeToxy improved text-based methods' performance and lessened keyword bias, paving the way for more robust audio abuse detection. Building on this work, Facebook AI extended on DeToxy by creating MuTox \cite{costa2024mutox}, one of the first large-scale multilingual audio datasets for toxicity detection. MuTox includes 20,000 audio utterances in English and Spanish and 4,000 utterances across 19 other languages. However, it is worth noting that this dataset only includes three Indian languages: Bengali, Hindi, and Urdu.

The first Bengali Audio Abuse dataset was introduced by \cite{rahut2020bengali}, where Transfer Learning was applied to extract features from 960 voice recordings of native Bengali speakers. Following this, ADIMA \cite{gupta2022} introduced the first Indian audio abuse dataset, comprising abusive audio clips in 10 Indian languages. This work sought to democratize audio-based content moderation in Indic languages through quantitative experiments conducted in both monolingual and cross-lingual zero-shot settings. A follow-up study by \cite{sharon2022} explored a multi-modal approach to improve abuse detection in multilingual settings. Additionally, \cite{spiesberger23_interspeech} demonstrated that abuse detection could be effectively performed using only acoustic and prosodic features on the ADIMA dataset, thereby avoiding the need for transcriptions.

\citet{sharon2024} leverage well-established Natural Language Processing techniques for abuse detection and introduce a cascaded model that combines an ASR system with textual keyword spotting and compares it with an end-to-end model utilizing audio-level feature embeddings and neural classifiers. A two-step process for abuse detection by first transcribing spoken audio into text using ASR systems followed by natural language processing-based methods was explored by \cite{sharon2022}. While this approach captured semantic information, it missed important audio cues such as pitch, volume, tone, and emotions, which are crucial in detecting abusive behaviour, as abusive speech often involves anger, agitation, or loudness \cite{rana2022emotionbasedhatespeech, plaza2021multi}.

\subsection{Few-Shot Learning and Meta-Learning}

Few-shot learning (FSL) is particularly significant in low-resource settings where data scarcity is a major challenge. Model-agnostic Meta-learning (MAML) has emerged as a powerful method within this domain. \citet{Singh2022Improved} proposed a MAML-based Low Resource ASR methodology using a multi-step loss (MSL) approach, which significantly improved the stability and accuracy of low-resource speech recognition systems compared to the traditional MAML approach. \citet{Gu2018MetaLearning} demonstrated the effectiveness of MAML in low-resource scenarios for Neural Machine Translation, significantly outperforming multilingual transfer learning methods on the Romanian-English WMT'16 dataset with only limited translated words. \citet{xia2021metaxl} proposed MetaXL, a method that effectively transforms representations from auxiliary languages to target languages, enhancing cross-lingual learning in tasks such as sentiment analysis and named entity recognition. 

\subsection{Automatic Speech Recognition}

Meta-learning approaches have also shown promise in automatic speech recognition. \citet{Hsu2019MetaLF} proposed MetaASR, which significantly outperformed the state-of-the-art multitask pretraining approach across various target languages with different combinations of pretraining languages. Furthermore, \cite{Conneau2020UnsupervisedCR} introduced XLSR, which learns cross-lingual speech representations by pretraining a single model from raw speech waveforms in multiple languages, enabling a single multilingual speech recognition model that is competitive with strong individual models.

\citet{Hou2021MetaAdapterEC} explored the combination of adapter modules with meta-learning algorithms to achieve high ASR performance in low-resource settings while improving parameter efficiency. In another study, \cite{Hou2021ExploitingAF} proposed SimAdapter, a novel algorithm for learning knowledge from adapters for cross-lingual speech adaptation, showing that these approaches can be integrated to achieve significant performance improvements, including a relative Word Error Rate (WER) reduction of up to 3.55\%.

While these works have highlighted the progress of various tasks like Abuse Detection, Machine Translation, Automatic Speech Recognition etc using Meta-Learning techniques in low-resource settings, there is an avenue for using Meta-Learning for Audio Abuse detection. This work contributes to the research gap and serves as a foundation for Audio Abuse Detection in Low Resource Languages, especially in Indian Languages, employing Few-Shot Learning and Pre-Trained Audio Representations.
\section{Methodologies} \label{methods}

Representations that are effective across general audio tasks, capture multiple robust features of the input sound thereby using these learned embeddings for classification tasks like Music Information Retrieval, Industrial Sound Analysis, etc \cite{niizumi2022byol, grollmisch2021analyzing}, we propose our method. Building on these studies, we employed a Model-Agnostic Meta-Learning (MAML) approach \cite{finn2017maml} to develop a few-shot classifier for cross-lingual audio abuse detection using pre-trained audio features. This approach leverages the adaptability of meta-learning to handle the complexities of low-resource and multilingual settings, ensuring improved performance in audio-based abusive content moderation. Additionally, we also perform a feature study of abusive language in the 10 languages with the best-performing normalised feature set that has the best classification accuracy. 

\subsection{Pre-Trained Audio Feature Extractions}

Features from Pre-Trained Audio Models were used for few-shot classification using MAML. We employed the CLSRIL-23 variant of Wav2Vec \cite{gupta2021clsril}, which is a self-supervised learning-based audio pre-trained model that learns cross-lingual speech representations from raw audio across 23 Indic languages. It is built on top of Wav2Vec 2.0 \cite{baevski2020wav2vec} and solved by training a contrastive task over masked latent speech representations and jointly learning the quantization of latents shared across all languages. We also used Whisper \cite{radford2022robust}, a pre-trained model for automatic speech recognition (ASR) and speech translation. Trained on 680k hours of labelled data, Whisper models demonstrated a strong ability to generalise to many datasets and domains without the need for fine-tuning.  

These extracted features were then normalised using the two methods:\\
\noindent \textbf{Temporal Mean: } This process involves computing the mean of the vectors along the temporal dimension for each tensor.
\begin{equation} \label{eq:temporal}
    V_i[j] = \frac{1}{x_i} \sum_{t=1}^{x_i} T_i[1, t, j]
\end{equation}

where \( x_i \) is the temporal length for the \( i \)$-th$ tensor and \( j \) ranges from \( 1 \) to the size of the feature dimension.

\noindent \textbf{L2-Norm: } This process involves computing the Euclidean Norm of the features for each tensor along the temporal dimension and then normalizing the features using the norm. After normalization, the mean vector is computed similarly to the previous function.
\begin{equation}\label{eq:l2-norm_1}
    f_{i,t}^{\text{norm}} = \frac{T_i[1, t, :]}{\sqrt{\sum_{j=1}^{768} (T_i[1, t, j])^2}}    
\end{equation}

\noindent The \( j \)$-th$ element of the mean vector \( V_i \) is:
\begin{equation} \label{eq:l2-norm_2}
    V_i[j] = \frac{1}{x_i} \sum_{t=1}^{x_i} f_{i,t}^{\text{norm}}[j]
\end{equation}

where \( x_i \) is the temporal length for the \( i \)$-th$ tensor and \( j \) ranges from \( 1 \) to the size of the feature dimension.

\subsection{Model Agnostic Meta-Learning (MAML)} \label{sec:maml}

Meta Agnostic Meta Learning is a technique introduced by \cite{finn2017maml}. The goal of few-shot meta-learning is to train a model that can quickly adapt to a new task using only a few data points and training iterations. For this, the model or learner is trained during a meta-learning phase on a set of tasks, such that the trained model can quickly adapt to new tasks using only a small number of examples or trials. In effect, the meta-learning problem treats entire tasks as training examples.

For our few-shot learning setup, we perform stratified sampling of $k$ samples per class for each language. Formally, let $\mathcal{D}_l = \{(x_i, y_i)\}_{i=1}^{N_l}$ denote the dataset of audio samples for language $l$, where $x_i$ represents the feature vector and $y_i$ the class label (abusive or non-abusive). For a given $k$-shot scenario, we construct a support set $\mathcal{S}_l \subseteq \mathcal{D}_l$ such that:
\begin{equation} \label{eq:set}
    \mathcal{S}_l = \{ (x_{i_1}, y_{i_1}), (x_{i_2}, y_{i_2}), \ldots, (x_{i_k}, y_{i_k}) \}
\end{equation}

where $y_{i_j} \in \{ \text{abusive}, \text{non-abusive} \}$ and each class is represented equally within the support set. Specifically, for a $k$-shot learning task with $k=2$, $\mathcal{S}_l$ contains one abusive and one non-abusive sample per language. 

Assuming there are $L$ languages, the total number of samples for the $k$-shot scenario is:
\begin{equation} \label{eq:sample_size}
    |\mathcal{S}| = k \times L
\end{equation}

where $\mathcal{S}$ represents the combined support sets across all languages. For instance, in a 50-shot scenario across 10 languages, this results in $50 \times 10 = 500$ samples.

\subsection{Cross-Lingual Training and Testing}

The few-shot model is trained using a cross-lingual approach, which is key to ensuring the model's ability to generalise across different languages. During training, the model is exposed to data from all $L$ languages, so that learning from the pre-trained representations captures the nuances of abusive and non-abusive speech across different contexts and languages. This cross-lingual training strategy enables the model to recognize similarities in abusive language across Indian languages, which is essential for achieving strong performance in low-resource scenarios where data for individual languages is limited. By leveraging a cross-lingual setting, the model can better generalize and identify abusive patterns even when specific language data is sparse. The testing phase involves evaluating the model's performance on individual languages and assessing its ability to adapt and accurately classify audio samples in a language-specific context after having been trained on a diverse set of languages. This setup simulates real-world conditions where a model, trained on data from multiple languages, must be capable of quickly adapting to and performing well on new languages with limited labelled examples. The effectiveness of this approach is demonstrated by the model's performance across various shot settings, which we analyze in detail in subsequent sections.

\subsection{Feature Study}

Observing the best-performing normalised feature set, this study aims to understand the specific characteristics of the features that contributed to improved classification accuracy in our few-shot learning setup. For the study, L2-Norm feature normalisation with Whisper was selected based on its superior performance in its accuracy scores as presented in Figure \ref{fig:l2normwhisper}. We performed a feature study by plotting the tSNE projection of the features to 2 dimensions and performing a visual analysis and an outlier analysis.
\begin{table}[bt]
    \centering
    \small
    \begin{tabular}{l|cccc|c}
        \hline
        \multicolumn{1}{c|}{\multirow{2}{*}{\textbf{Language}}} & \multicolumn{2}{c|}{\textbf{Abusive}} & \multicolumn{2}{c|}{\textbf{Non-Abusive}} & \multicolumn{1}{l}{\multirow{2}{*}{\textbf{Total}}} \\ \cline{2-5}
        \multicolumn{1}{c|}{} & Train & \multicolumn{1}{c|}{Test} & Train & Test & \multicolumn{1}{l}{} \\ \hline
        Bengali & 394 & \multicolumn{1}{c|}{148} & 428 & 222 & 1192 \\
        Bhojpuri & 253 & \multicolumn{1}{c|}{122} & 506 & 214 & 1095 \\
        Gujarati & 516 & \multicolumn{1}{c|}{255} & 301 & 107 & 1179 \\
        Haryanvi & 419 & \multicolumn{1}{c|}{193} & 399 & 173 & 1184 \\
        Hindi & 449 & \multicolumn{1}{c|}{186} & 373 & 183 & 1191 \\
        Kannada & 530 & \multicolumn{1}{c|}{243} & 289 & 126 & 1188 \\
        Malayalam & 582 & \multicolumn{1}{c|}{257} & 237 & 115 & 1191 \\
        Odia & 491 & \multicolumn{1}{c|}{209} & 323 & 156 & 1179 \\
        Punjabi & 405 & \multicolumn{1}{c|}{176} & 413 & 191 & 1185 \\
        Tamil & 572 & \multicolumn{1}{c|}{267} & 248 & 104 & 1191 \\ \hline
        \textbf{Total} & \multicolumn{1}{r}{4611} & \multicolumn{1}{r|}{2056} & \multicolumn{1}{r}{3517} & \multicolumn{1}{r|}{1591} & \multicolumn{1}{r}{11775}\\
        \hline
    \end{tabular}
    \caption{ADIMA  Dataset distribution across languages and classes. Train and Test being the ones provided by authors.}
    \label{tab:audio-stats}
\end{table}
\begin{figure*}[htbp]
     \centering
     \begin{subfigure}[htbp]{0.47\textwidth}
         \centering
         \includegraphics[width=\textwidth]{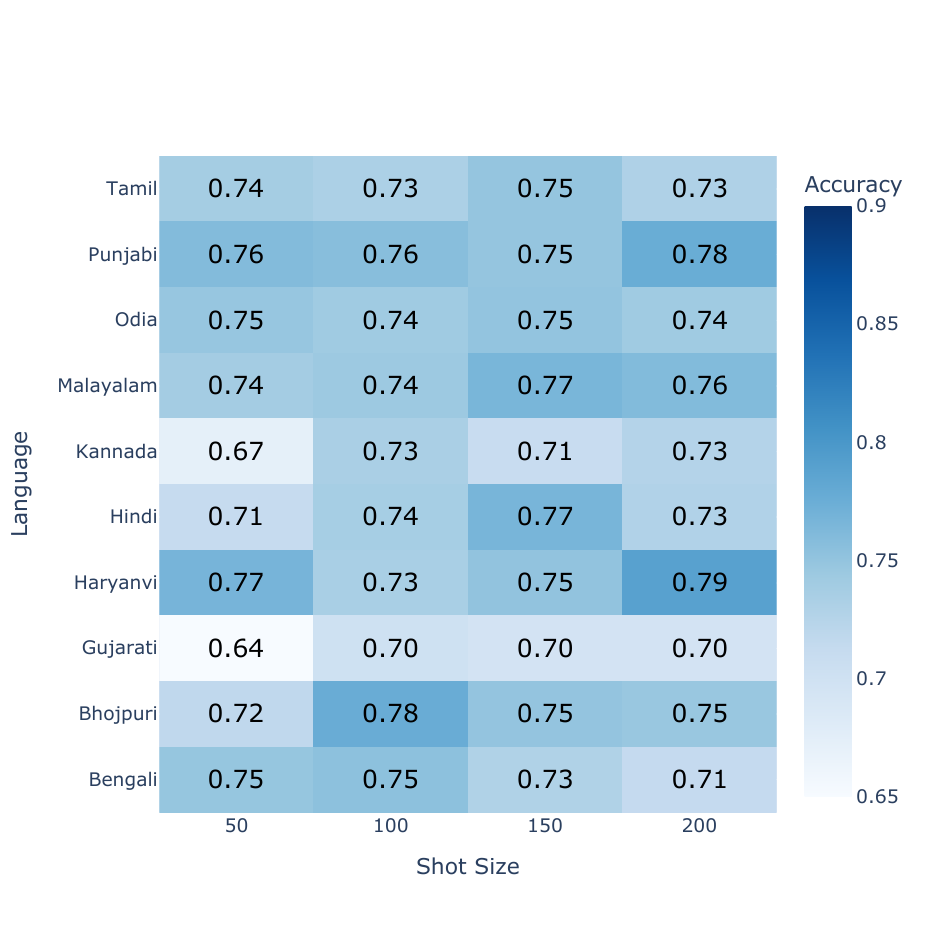}
         \vspace{-0.9cm}
         \caption{Temporal Mean Wav2Vec}
         \label{fig:tempwav}
     \end{subfigure}
     \hfill
     \begin{subfigure}[htbp]{0.47\textwidth}
         \centering
         \includegraphics[width=\textwidth]{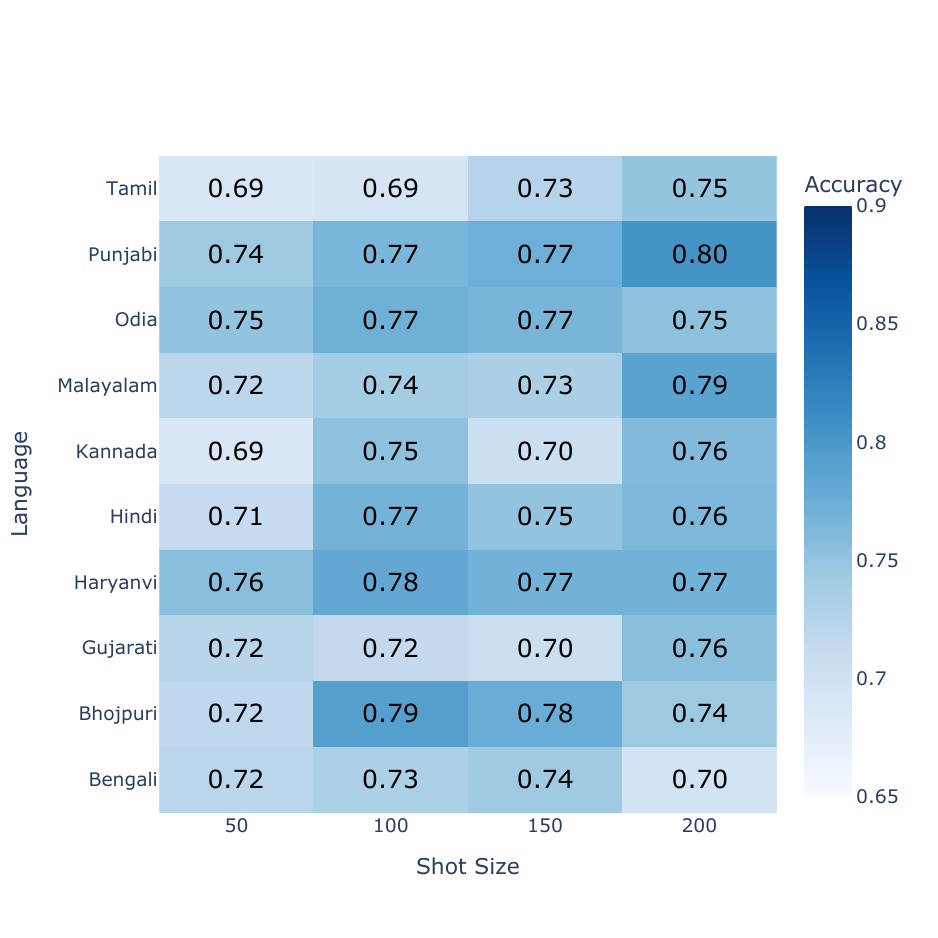}
         \vspace{-0.9cm}
         \caption{Temporal Mean Whisper}
         \label{fig:tempwhisper}
     \end{subfigure}
     \vspace{-0.1cm}
     \caption{\textbf{Temporal Mean}: Few Shot Accuracies in 50, 100, 150 and 200 shot cases}
     \label{fig:temporal-mean-results}
\end{figure*}
\section{Experiments} \label{experiments}

\subsection{Dataset}

To carry out our study, we employed the ADIMA dataset \cite{gupta2022} by ShareChat. It contains 11,775 audio clips, sourced from real-life conversations, in 10 Indian Languages annotated for a binary classification task. It is an evenly distributed dataset comprised of 5,108 abusive and 6,667 non-abusive samples from 6,446 unique users. Table~\ref{tab:audio-stats} presents the distribution of samples across the classification category and across 10 languages where the average and median number of samples are 1177.5 and 1186.5 respectively.  We can observe that Bhojpuri has the least number of samples, but the maximum number of abusive samples, whereas Bengali has the highest number of audio samples. Another observation is the lesser amount of abusive samples Malayalam and Tamil have compared to non-abusive samples.

\subsection{Feature Extraction}

The code for extracting features for all the audio clips was with PyTorch \cite{pytorch} and HuggingFace libraries. In this task, we compared two pre-trained audio models: Whisper \cite{radford2022robust} and Wav2Vec \cite{baevski2020wav2vec}, specifically, the \texttt{whisper-large} variant for Whisper and the \texttt{CLSRIL-23} variant of Wav2Vec \cite{gupta2021clsril} were used for extracting pre-trained features. Firstly, embeddings were extracted by passing raw audio files through these models. The embeddings were then feature-normalised to generate decision-level features \cite{wang2023multimodal} for our task in two ways: Temporal Mean (equation~\ref{eq:temporal}) and L2-Norm (equation~\ref{eq:l2-norm_1}) to generate two different feature sets.

\subsection{Few-Shot Experimental Setup}

We employed the Model-Agnostic Meta-Learning (MAML) algorithm \cite{finn2017maml} to address the challenge of few-shot cross-lingual audio abuse detection. The few-shot learning methodology is particularly suited to scenarios where only a limited number of labelled examples are available for each language, ensuring efficient and effective learning from small datasets. We utilized stratified sampling \cite{mitchell1996consequences} to sample audio clips for the few-shot task ensuring that the proportion of abusive and non-abusive samples remained balanced across languages.  As discussed in Section~\ref{sec:maml}, for a shot size "$k$", $k$-samples are chosen in the 10 languages, thereby making the number of samples for cross-lingual training 10 x $k$-samples (refer equation~\ref{eq:sample_size}). To evaluate the model’s performance across different few-shot settings, we conducted experiments with varying shot sizes: 50, 100, 150, and 200. These shot sizes were selected to investigate the impact of sample size on model performance, particularly in scenarios where the number of available samples is less than half of the average number of audio clips per language, which is approximately 1177.5 (refer to Table~\ref{tab:audio-stats}). By exploring these shot sizes, we aimed to understand how well the model could generalize to unseen examples with limited training data, a key concern in real-world applications where extensive labelled data is often unavailable. The data was split into training and testing sets based on the splits provided by \cite{gupta2022}.

\subsection{Model Architecture and Training}

The learner model utilized in our experiments is an Artificial Neural Network (ANN) consisting of three fully connected layers. The network architecture was designed as follows: an input layer with a size corresponding to the dimension of the extracted feature vectors (1024 for Whisper and 768 for Wav2Vec), followed by hidden layers with sizes 256 and 128 respectively, and a final output layer with size 2, with leaky ReLu for non-linearity, corresponding to the binary classification task. The output layer utilized a softmax activation function to convert the raw logits into probabilities for each class, determining whether an audio clip was abusive or non-abusive. Training of the learner model was done using the Adam optimizer \cite{kingma2017adam}. We employed a task-specific learning rate and a meta-learning rate of 0.001, both of which were managed by a linear learning rate scheduler with default parameters as provided by the PyTorch optimizer library. The model was trained with a batch size of 128 and for a total of 150 epochs, based on repeated testing, which provided a good balance between training time and model performance. 
\begin{figure*}[ht]
     \centering
     \begin{subfigure}[htbp]{0.47\textwidth}
         \centering
         \includegraphics[width=\textwidth]{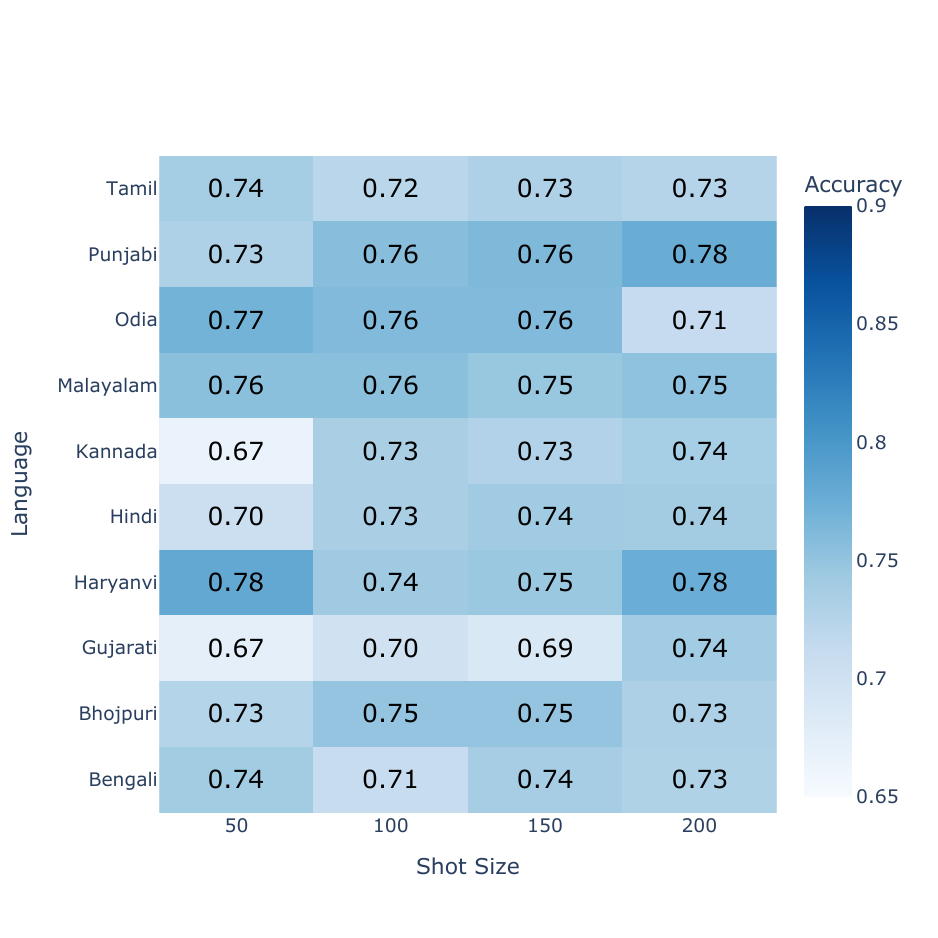}
                  \vspace{-0.9cm}
         \caption{L2-Norm Wav2Vec}
         \label{fig:l2normwav}
     \end{subfigure}
     \hfill
     \begin{subfigure}[htbp]{0.47\textwidth}
         \centering
         \includegraphics[width=\textwidth]{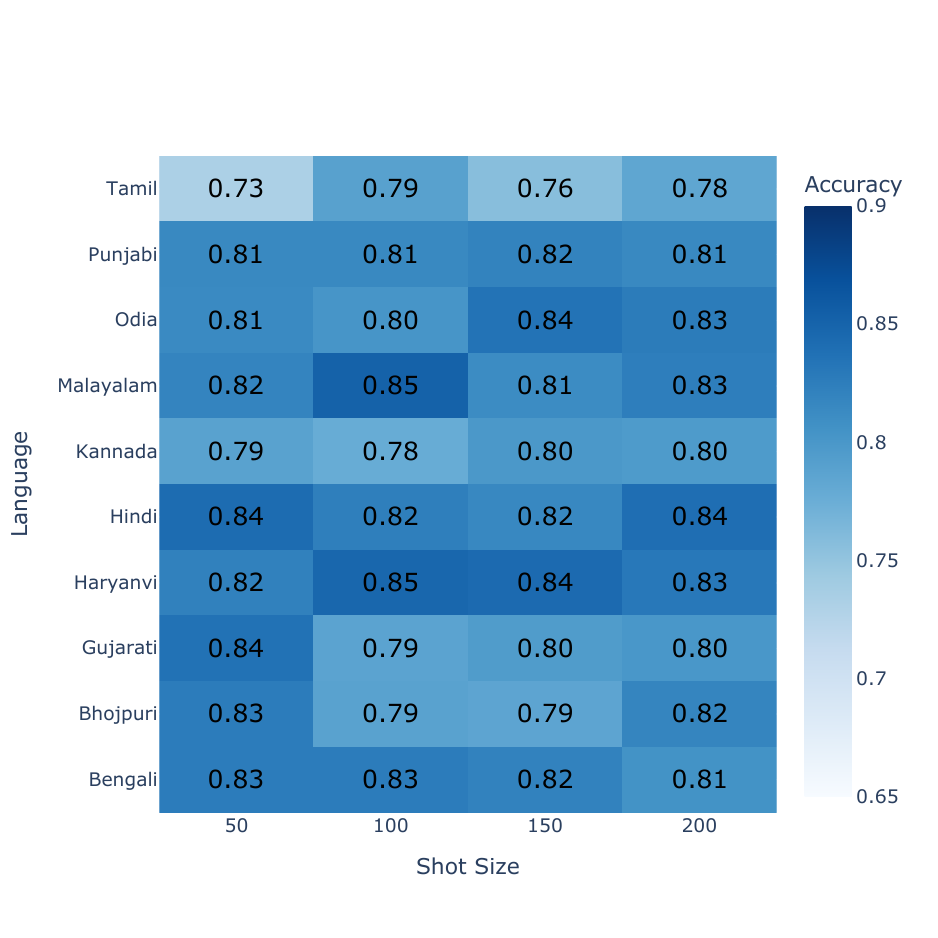}
                  \vspace{-0.9cm}
         \caption{L2-Norm Whisper}
         \label{fig:l2normwhisper}
     \end{subfigure}
              \vspace{-0.2cm}
     \caption{\textbf{L2-Norm}: Few Shot Accuracies in 50, 100, 150 and 200 shot cases}
     \label{fig:l2-norm-results}
\end{figure*}
\section{Results} \label{result}

\subsection{Classification Results}\label{res:class}

We present accuracy scores from both feature settings in 4 shot settings [50, 100, 150, 200] as heat-maps in Figures~\ref{fig:temporal-mean-results} and~\ref{fig:l2-norm-results}. More Detailed results with macro-F1 scores are provided in the Appendix~\ref{appendix}. An aggregate macro-f1 score table with the baseline from the Dataset paper \cite{gupta2022} is also presented in the Appendix in Table~\ref{tab:aggregate}.

It is evident that Whisper with the L2-Norm feature normalisation has consistently better scores across languages, with no top accuracy scores in the 200-shot scenario and most of the best-performing accuracy scores are in the 50 and 100-shot settings. Comparing normalisation settings, we can observe that L2-Norm has much better classification performance compared to the temporal mean normalisation setting. For most languages, L2-Norm offers higher accuracy and macro-F1 scores compared to Temporal Mean. For Bhojpuri with Whisper, L2-Norm gives significantly better accuracy (82.75\% at 50 shots) compared to Temporal Mean (79.17\% at 100 shots). Some other results that are evident are the not-so-consistent performance of Tamil and Kannada, especially Tamil, with its highest Accuracy scores being 74.93\% in the Temporal Mean Normalisation setting of Wav2Vec and 78.98\% in the L2-Norm Mean Normalisation of Whisper. Across both models and normalization methods, F1 scores generally tracked closely with accuracy. 
Languages like Haryanvi, Punjabi, and Odia generally perform better than other languages across both models and normalization strategies. For Whisper with L2-Norm, Haryanvi has strong accuracy (84.7\% at 100 shots), and similarly for Punjabi and Odia. Gujarati, Kannada, and Tamil show lower accuracy and F1 scores overall compared to others, regardless of the model or normalization method. An interesting observation is the abuse detection accuracy score of Malayalam with 85.22\% accuracy in the L2-Norm feature normalisation setting with Whisper features. Given that Tamil, Kannada and Malayalam are of the Dravidian Language family \cite{Srivatsa2017The}, it is interesting to observe Malayalam's superior accuracy scores compared to its other Language family counterparts in the cross-lingual setting. 

\subsection{Pre-Trained Feature Study of Abusive Language}\label{sec:PTA}

\begin{figure*}[ht]
    \centering
    \includegraphics[width=1\linewidth]{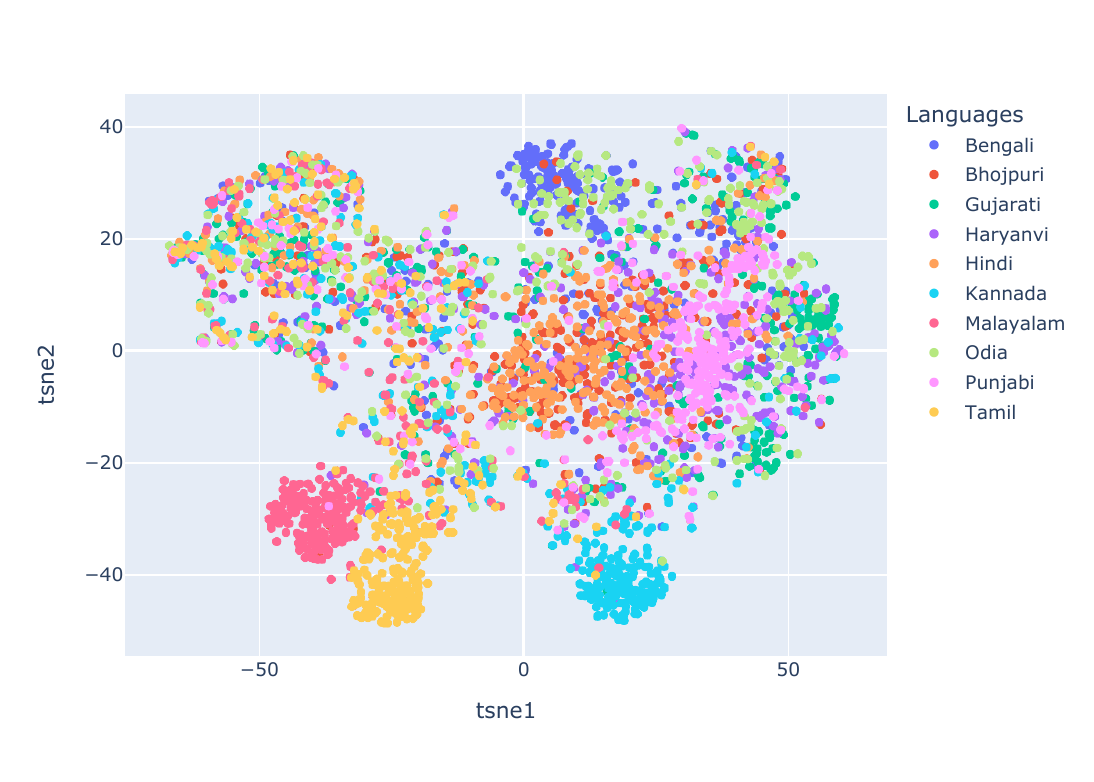}
    \vspace{-1.4cm}
    \caption{tSNE plot of L2-Norm Feature normalisation of Whisper Features extracted from the ADIMA dataset}
    \label{fig:Whisper-L2-tSNE}
\end{figure*}

Pre-Trained audio representations offer a powerful perspective for studying abusive language, especially in Low-Resource contexts where labelled data can be scarce or imbalanced. Models like Whisper, employing vast amounts of training data to learn robust audio feature representations enable transfers to downstream tasks like abuse detection. This approach allows us to bypass the need for large annotated datasets and explore the underlying audio characteristics that distinguish abusive language across diverse linguistic groups. To understand the improved performance of L2-norm feature normalization in conjunction with Whisper’s ability to perform cross-lingual abuse detection, as demonstrated by its superior performance in Section~\ref{res:class}, we visualized the extracted features by plotting them in 2D with t-SNE and observed the language clustering by performing a visual study of the features.

As shown in Figure~\ref{fig:Whisper-L2-tSNE}, three distinct clusters emerge for the Dravidian languages (Kannada, Malayalam, and Tamil) present in the ADIMA dataset. These clusters are separated from others, with Tamil and Malayalam appearing closer together, which is expected given their linguistic similarities. Malayalam shares significant grammatical and literary ties with Tamil, particularly from the historical period spanning late Old Tamil to early Middle Tamil \cite{G2022Family, Menon1990Some}. In contrast, the Indo-Aryan language family \cite{Jain2007The} forms a denser and more overlapping cluster, encompassing languages such as Bengali, Bhojpuri, Gujarati, Haryanvi, Hindi, Odia, and Punjabi. Within this cluster, languages like Bengali and Odia appear grouped, while Hindi and Punjabi show a similar pattern clustered close together.
The distinct clusters observed in the t-SNE plot suggest that Whisper’s audio features capture language-specific patterns effectively in the L2-Norm feature normalization. This finding is promising for cross-lingual classification tasks, as it indicates the preservation of key phonetic characteristics across languages. For instance, Tamil and Malayalam form tighter clusters, reflecting their clearer phonetic or acoustic features, which can aid classification. However, languages such as Bhojpuri or Haryanvi exhibit less defined boundaries, likely due to limited data and findings of them being dialects of Hindi \cite{Sinha2014Speech}, which results in noisier feature representations. These overlaps help support the study that Hindi-like languages are similar acoustically as well and provide insight that maybe language does play a part in cross-lingual audio abuse detection.
\section{Conclusion} \label{conclusion}
With the widespread adoption of social media for everyday communication, the need for effective content moderation has become critical to keeping malicious actors in check. While there has been extensive research on text-based moderation, safeguarding against abusive content in audio remains underexplored, particularly for low-resource languages. In this work, we propose a few-shot cross-lingual audio abuse detection method in low-resource languages, specifically focusing on Indian languages by employing the Model-Agnostic Meta-Learning (MAML) framework \cite{finn2017maml}. Our method addresses the challenge of detecting abusive audio content with limited training samples per language, a key issue in low-resource settings by benchmarking our approach on the ADIMA dataset \cite{gupta2022}, which provides binary-labeled audio clips for abuse detection in 10 Indian Languages, leveraging pre-trained audio representations from Wav2Vec and Whisper.

To provide a deeper understanding to enhance the performance of the few-shot classification, we investigated the impact of various feature normalization techniques, such as Temporal Mean Normalization and L2-Norm, applied to the extracted features from the pre-trained audio models and we also conducted a comparative study across these normalization methods and pre-trained models. To better understand the effect of feature normalization and to identify the best-performing model, we conduct a visual analysis of the learned audio representations by plotting the features using a 2D t-SNE plot. We were able to observe language similarities and have discussed why L2-Norm with Whisper Features performed well compared to other feature normalisation techniques and Pre-Trained Audio Representations (Section~\ref{sec:PTA}).

Our research contributes to the ongoing efforts to combat abusive content on social media platforms, particularly in the audio domain, by providing insights into meta-learning, feature normalization, and performance in low-resource language settings.

\section*{Limitations}

While this work explores cross-lingual few-shot audio abuse detection in the 10 languages ADIMA provides, we believe this methodology can also be expanded to other low-resource languages. Further research will be needed to assess efficiency. Exploring other Meta Learning Algorithms like ProtoMAML \cite{Triantafillou2020protomaml} and Contrastive Learning \cite{saeed2021contrastive} will also contribute to addressing these issues. While this is a cross-lingual abuse detection task, there are avenues for Mono-Lingual Experiments too for more specific languages and also with other languages for cross-lingual tasks. Whisper and Wav2Vec have been the only models that have been used but future works will involve exploring other pre-trained audio models like SeamlessM4T \cite{seamless2023} and different feature normalisation such as other L-N Normalisation, Weighted Averaging \cite{phukan2024collab} and more.

Since this work deals with Low Resource Languages in the Indian context, an important limitation is the absence of training data in other Indian Languages since Languages like Telugu and Marathi have been missed out, which are other major spoken languages with about 71 million people who speak Marathi as their native tongue. \cite{Garje2016Marathi} and Telugu with 82.7 million native speakers \cite{Jaswanth2022A}. Training data in these languages will add diversity and more languages to detect, improving variety and diversity, thereby being inclusive of all languages. Given the scarcity, we would also like to see the creation of curated datasets for offensive speech detection in other Low-Resource Languages, including ones from the Global South, in the audio modality.

\section*{Ethics Statement}

This study does not involve any personal or public data pointing to an individual or a group of individuals and thus does not break any ethical guidelines.
\bibliography{ref}
\appendix
\newpage
\onecolumn
\section{Appendix} \label{appendix}

\subsection{Accuracy Tables}

To improve readability and facilitate analysis of the extensive results, we have organized the accuracy and F1 scores for various shot sizes and normalization settings into dedicated tables here in the appendix section.

\begin{table*}[h]   
\small
\begin{adjustbox}{width=\textwidth}

\begin{tabular}{lcccccccc?cccccccc}
\hline
\multicolumn{1}{c}{} & \multicolumn{8}{c?}{\textbf{Temporal Mean}} & \multicolumn{8}{c}{\textbf{L2-Norm}} \\ \hline
\multicolumn{1}{c|}{Shot Size} & \multicolumn{2}{c|}{50} & \multicolumn{2}{c|}{100} & \multicolumn{2}{c|}{150} & \multicolumn{2}{c?}{200} & \multicolumn{2}{c|}{50} & \multicolumn{2}{c|}{100} & \multicolumn{2}{c|}{150} & \multicolumn{2}{c}{200} \\ \hline
\multicolumn{1}{l|}{Language} & Acc & \multicolumn{1}{c|}{F1} & Acc & \multicolumn{1}{c|}{F1} & Acc & \multicolumn{1}{c|}{F1} & Acc & F1 & Acc & \multicolumn{1}{c|}{F1} & Acc & \multicolumn{1}{c|}{F1} & Acc & \multicolumn{1}{c|}{F1} & Acc & F1 \\ \hline
\multicolumn{1}{l|}{Bengali} & 74.86 & \multicolumn{1}{c|}{74.3} & \textbf{75.41} & \multicolumn{1}{c|}{\textbf{74.57}} & 72.97 & \multicolumn{1}{c|}{72.38} & 71.35 & 70.73 & \textbf{74.05} & \multicolumn{1}{c|}{\textbf{73.57}} & 71.08 & \multicolumn{1}{c|}{70.61} & 73.78 & \multicolumn{1}{c|}{72.9} & 72.97 & 72.24 \\
\multicolumn{1}{l|}{Bhojpuri} & 71.73 & \multicolumn{1}{c|}{69.59} & \textbf{77.68} & \multicolumn{1}{c|}{\textbf{75.91}} & 75 & \multicolumn{1}{c|}{73.25} & 74.7 & 73.06 & 72.62 & \multicolumn{1}{c|}{70.97} & \textbf{75} & \multicolumn{1}{c|}{\textbf{73.58}} & 75 & \multicolumn{1}{c|}{72.97} & 73.21 & 71.04 \\
\multicolumn{1}{l|}{Gujarati} & 64.09 & \multicolumn{1}{c|}{62.61} & \textbf{70.17} & \multicolumn{1}{c|}{\textbf{68.29}} & 69.61 & \multicolumn{1}{c|}{66.76} & 69.61 & 67.61 & 67.4 & \multicolumn{1}{c|}{65.64} & 69.89 & \multicolumn{1}{c|}{67.35} & 68.78 & \multicolumn{1}{c|}{66.48} & \textbf{74.03} & \textbf{71.28} \\
\multicolumn{1}{l|}{Haryanvi} & 76.78 & \multicolumn{1}{c|}{76.68} & 73.5 & \multicolumn{1}{c|}{73.47} & 75.14 & \multicolumn{1}{c|}{75.02} & \textbf{78.96} & \textbf{78.83} & \textbf{78.14} & \multicolumn{1}{c|}{\textbf{78.1}} & 74.32 & \multicolumn{1}{c|}{74.27} & 74.59 & \multicolumn{1}{c|}{74.49} & 77.6 & 77.57 \\
\multicolumn{1}{l|}{Hindi} & 71.27 & \multicolumn{1}{c|}{71.25} & 73.71 & \multicolumn{1}{c|}{73.7} & \textbf{76.69} & \multicolumn{1}{c|}{\textbf{76.69}} & 72.9 & 72.88 & 70.46 & \multicolumn{1}{c|}{70.44} & 73.44 & \multicolumn{1}{c|}{73.43} & \textbf{74.25} & \multicolumn{1}{c|}{\textbf{74.23}} & 73.98 & 73.98 \\
\multicolumn{1}{l|}{Kannada} & 67.21 & \multicolumn{1}{c|}{66.44} & \textbf{73.44} & \multicolumn{1}{c|}{\textbf{72.36}} & 71 & \multicolumn{1}{c|}{69.42} & 72.63 & 71.06 & 66.67 & \multicolumn{1}{c|}{65.7} & 73.44 & \multicolumn{1}{c|}{71.88} & 72.9 & \multicolumn{1}{c|}{71.3} & \textbf{73.71} & \textbf{72.05} \\
\multicolumn{1}{l|}{Malayalam} & 73.92 & \multicolumn{1}{c|}{72.27} & 74.46 & \multicolumn{1}{c|}{72.52} & \textbf{76.61} & \multicolumn{1}{c|}{\textbf{74.67}} & 76.08 & 74 & \textbf{75.54} & \multicolumn{1}{c|}{\textbf{73.83}} & 75.54 & \multicolumn{1}{c|}{73.33} & 74.73 & \multicolumn{1}{c|}{72.68} & 75.27 & 73.58 \\
\multicolumn{1}{l|}{Odia} & 74.79 & \multicolumn{1}{c|}{74.16} & 74.25 & \multicolumn{1}{c|}{73.85} & \textbf{75.07} & \multicolumn{1}{c|}{\textbf{74.55}} & 74.25 & 73.77 & \textbf{76.99} & \multicolumn{1}{c|}{\textbf{76.28}} & 76.16 & \multicolumn{1}{c|}{75.59} & 76.16 & \multicolumn{1}{c|}{75.59} & 71.23 & 70.92 \\
\multicolumn{1}{l|}{Punjabi} & 76.02 & \multicolumn{1}{c|}{75.62} & 75.75 & \multicolumn{1}{c|}{75.21} & 74.93 & \multicolumn{1}{c|}{74.62} & \textbf{77.66} & \textbf{77.43} & 73.02 & \multicolumn{1}{c|}{72.51} & 75.75 & \multicolumn{1}{c|}{75.36} & 76.29 & \multicolumn{1}{c|}{76.01} & \textbf{77.66} & \textbf{77.4} \\
\multicolumn{1}{l|}{Tamil} & 73.85 & \multicolumn{1}{c|}{71.23} & 73.32 & \multicolumn{1}{c|}{70.64} & \textbf{74.93} & \multicolumn{1}{c|}{\textbf{72.11}} & 73.05 & 70.4 & \textbf{73.85} & \multicolumn{1}{c|}{\textbf{71.13}} & 72.24 & \multicolumn{1}{c|}{68.87} & 73.05 & \multicolumn{1}{c|}{70.4} & 72.51 & 69.8 \\ \hline
\end{tabular}

\end{adjustbox}
\caption{Few-shot Classification Results for Wav2Vec \\{Acc: Accuracy, F1: Macro F1-Score}}
\label{tab:Wav2Vec-complete}
\end{table*}
\begin{table*}[h]
\small
\begin{adjustbox}{width=\textwidth}

\begin{tabular}{lcccccccc?cccccccc}
\hline
\multicolumn{1}{c}{} & \multicolumn{8}{c?}{\textbf{Temporal Mean}} & \multicolumn{8}{c}{\textbf{L2-Norm}} \\ \hline
\multicolumn{1}{c|}{Shot Size} & \multicolumn{2}{c|}{50} & \multicolumn{2}{c|}{100} & \multicolumn{2}{c|}{150} & \multicolumn{2}{c?}{200} & \multicolumn{2}{c|}{50} & \multicolumn{2}{c|}{100} & \multicolumn{2}{c|}{150} & \multicolumn{2}{c}{200} \\ \hline
\multicolumn{1}{l|}{Language} & Acc & \multicolumn{1}{c|}{F1} & Acc & \multicolumn{1}{c|}{F1} & Acc & \multicolumn{1}{c|}{F1} & Acc & F1 & Acc & \multicolumn{1}{c|}{F1} & Acc & \multicolumn{1}{c|}{F1} & Acc & \multicolumn{1}{c|}{F1} & Acc & F1 \\ \hline
\multicolumn{1}{l|}{Bengali} & 72.16 & \multicolumn{1}{c|}{71.74} & 73.24 & \multicolumn{1}{c|}{73} & \textbf{74.32} & \multicolumn{1}{c|}{\textbf{73.65}} & 69.73 & 69.66 & \textbf{82.7} & \multicolumn{1}{c|}{\textbf{82.45}} & 82.7 & \multicolumn{1}{c|}{82.33} & 82.16 & \multicolumn{1}{c|}{81.91} & 80.54 & 79.9 \\
\multicolumn{1}{l|}{Bhojpuri} & 71.73 & \multicolumn{1}{c|}{70.63} & \textbf{79.17} & \multicolumn{1}{c|}{\textbf{77.85}} & 77.68 & \multicolumn{1}{c|}{75.56} & 74.4 & 73.62 & \textbf{82.74} & \multicolumn{1}{c|}{\textbf{81.34}} & 78.87 & \multicolumn{1}{c|}{77.12} & 78.57 & \multicolumn{1}{c|}{76.58} & 81.85 & 80.19 \\
\multicolumn{1}{l|}{Gujarati} & 72.38 & \multicolumn{1}{c|}{69.09} & 71.55 & \multicolumn{1}{c|}{68.23} & 70.44 & \multicolumn{1}{c|}{68.26} & \textbf{75.69} & \textbf{72.09} & \textbf{83.7} & \multicolumn{1}{c|}{\textbf{81.73}} & 78.73 & \multicolumn{1}{c|}{76.43} & 79.56 & \multicolumn{1}{c|}{77.56} & 80.11 & 77.75 \\
\multicolumn{1}{l|}{Haryanvi} & 75.68 & \multicolumn{1}{c|}{75.67} & \textbf{78.14} & \multicolumn{1}{c|}{\textbf{78.14}} & 77.05 & \multicolumn{1}{c|}{76.99} & 77.05 & 76.93 & 82.24 & \multicolumn{1}{c|}{82.21} & \textbf{84.7} & \multicolumn{1}{c|}{\textbf{84.69}} & 84.43 & \multicolumn{1}{c|}{84.4} & 83.06 & 83.04 \\
\multicolumn{1}{l|}{Hindi} & 71.27 & \multicolumn{1}{c|}{71.26} & \textbf{76.96} & \multicolumn{1}{c|}{\textbf{76.94}} & 75.07 & \multicolumn{1}{c|}{75.03} & 76.42 & 76.2 & \textbf{84.28} & \multicolumn{1}{c|}{\textbf{84.26}} & 82.38 & \multicolumn{1}{c|}{82.37} & 81.57 & \multicolumn{1}{c|}{81.53} & 84.01 & 84 \\
\multicolumn{1}{l|}{Kannada} & 68.83 & \multicolumn{1}{c|}{68.2} & 75.34 & \multicolumn{1}{c|}{73.21} & 70.46 & \multicolumn{1}{c|}{70.26} & \textbf{76.15} & \textbf{74.14} & 78.86 & \multicolumn{1}{c|}{77.95} & 77.78 & \multicolumn{1}{c|}{76.77} & \textbf{79.95} & \multicolumn{1}{c|}{\textbf{78.77}} & 79.67 & 78.67 \\
\multicolumn{1}{l|}{Malayalam} & 72.04 & \multicolumn{1}{c|}{68.92} & 73.92 & \multicolumn{1}{c|}{70.48} & 73.39 & \multicolumn{1}{c|}{71.36} & \textbf{78.76} & \textbf{75.08} & 81.99 & \multicolumn{1}{c|}{80.23} & \textbf{85.22} & \multicolumn{1}{c|}{\textbf{83.33}} & 81.18 & \multicolumn{1}{c|}{78.83} & 82.53 & 80.3 \\
\multicolumn{1}{l|}{Odia} & 75.07 & \multicolumn{1}{c|}{74.38} & \textbf{77.26} & \multicolumn{1}{c|}{\textbf{76.58}} & 76.71 & \multicolumn{1}{c|}{76.48} & 75.34 & 74.07 & 81.37 & \multicolumn{1}{c|}{80.8} & 80.27 & \multicolumn{1}{c|}{79.78} & \textbf{83.56} & \multicolumn{1}{c|}{\textbf{83.18}} & 82.74 & 82.35 \\
\multicolumn{1}{l|}{Punjabi} & 74.39 & \multicolumn{1}{c|}{74.2} & 76.57 & \multicolumn{1}{c|}{76.38} & 77.38 & \multicolumn{1}{c|}{77.11} & \textbf{80.38} & \textbf{80.36} & 81.47 & \multicolumn{1}{c|}{81.3} & 81.47 & \multicolumn{1}{c|}{81.21} & \textbf{82.02} & \multicolumn{1}{c|}{\textbf{81.87}} & 81.47 & 81.32 \\
\multicolumn{1}{l|}{Tamil} & 69 & \multicolumn{1}{c|}{65.24} & 69.27 & \multicolumn{1}{c|}{65.34} & 72.51 & \multicolumn{1}{c|}{70.31} & \textbf{74.93} & \textbf{71.42} & 73.32 & \multicolumn{1}{c|}{70.74} & \textbf{78.98} & \multicolumn{1}{c|}{\textbf{75.88}} & 75.74 & \multicolumn{1}{c|}{72.85} & 78.44 & 76.14 \\ \hline
\end{tabular}

\end{adjustbox}
\caption{Few-shot Classification Results for Whisper \\{Acc: Accuracy, F1: Macro F1-Score}}
\label{tab:Whisper-complete}
\end{table*}

\subsection{Aggregate Scores Table}

This section and the presented table is with regard to a reviewer's concern about the need to compare with a baseline. 
The original Dataset paper for ADIMA \cite{gupta2022} presents Macro F1 Scores by training a Zero-shot model on the source language and evaluating the performance on the target language using CLSRIL-23 \cite{gupta2021clsril} and Max-Pooling. 

\begin{table}[h]
\centering
\begin{tabular}{l|ll}
\textbf{Language}  & \texttt{ADIMA}         & Ours           \\ \hline
Bengali   & 79.1          & \textbf{82.45} \\
Bhojpuri  & -             & \textbf{81.34} \\
Gujarati  & -             & \textbf{81.73} \\
Haryanvi  & -             & \textbf{84.69} \\
Hindi     & 80.7          & \textbf{84.26} \\
Kannada   & 78.4          & \textbf{78.77} \\
Malayalam & -             & \textbf{83.33} \\
Odia      & -             & \textbf{83.18} \\
Punjabi   & \textbf{83.4} & 81.87          \\
Tamil     & 75.2          & \textbf{75.88}
\end{tabular}
\caption{Baseline ADIMA vs Aggregate Macro F1 Scores for comparison}
\label{tab:aggregate}
\end{table}

In table~\ref{tab:aggregate}, we present a comparison of our best scores versus their approach for better clarity.
\end{document}